\documentclass[letterpaper, 10 pt, conference]{ieeeconf}
\IEEEoverridecommandlockouts
\usepackage{tablefootnote}

\def\usenatbib{1}
\if\usenatbib1
\makeatletter
\let\NAT@parse\undefined
\makeatother
\usepackage[numbers]{natbib}
\else
\usepackage{cite}
\fi

\usepackage[hyphens]{url}
\usepackage{graphicx}
\usepackage{xcolor}
\usepackage{hyperref}
\usepackage{amsmath}
\usepackage{amssymb}
\usepackage{algorithm}
\usepackage{algorithmic}
\usepackage{booktabs}
\usepackage{cancel}
\usepackage[
separate-uncertainty=true,
detect-weight,
table-number-alignment = center,
table-text-alignment = center,
table-format=2.2(4),
table-sign-mantissa,
]{siunitx}
\usepackage{cleveref}
\usepackage[T1]{fontenc}
\usepackage{todonotes}
\usepackage{pgfplots}

\pgfplotsset{compat=newest}
\DeclareMathOperator*{\argmax}{arg\,max}
\newcommand{\cI}{\mathcal{I}}
\newcommand{\cS}{\mathcal{S}}
\newcommand{\cA}{\mathcal{A}}
\newcommand{\cO}{\mathcal{O}}
\newcommand{\cP}{\mathcal{P}}
\newcommand{\cR}{\mathcal{R}}

\usepackage{bbm}

\begin{document}

\title{\LARGE \bf
Scalable Multi-Agent Reinforcement Learning for Warehouse Logistics with Robotic and Human Co-Workers
}

\author{
\authorblockN{
Aleksandar Krnjaic\authorrefmark{1},
Raul D. Steleac\authorrefmark{1}\authorrefmark{2},
Jonathan D. Thomas\authorrefmark{1}\authorrefmark{2},
Georgios Papoudakis\authorrefmark{1}\authorrefmark{2}, \\
Lukas Schäfer\authorrefmark{1}\authorrefmark{2},
Andrew Wing Keung To\authorrefmark{1},
Kuan-Ho Lao\authorrefmark{1},
Murat Cubuktepe\authorrefmark{1},\\
Matthew Haley\authorrefmark{1},
Peter Börsting\authorrefmark{1},
Stefano V. Albrecht\authorrefmark{1}\authorrefmark{2}}
\vspace{0.5em}
\authorblockA{
aleks.krnjaic@dematic.com,
raul.steleac@ed.ac.uk,
jonathan.d.thomas8@gmail.com,
g.papoudakis@ed.ac.uk, \\
l.schaefer@ed.ac.uk,
andrew.to@dematic.com,
kuanho.lao@dematic.com,
murat.cubuktepe@dematic.com, \\
matthew.haley@dematic.com,
peter.boersting@dematic.com,
s.albrecht@ed.ac.uk
}
\vspace{0.5em}
\authorblockA{\authorrefmark{1}Dematic,
\authorrefmark{2}University of Edinburgh, UK}
\thanks{\hspace{-9pt}Dematic is a multinational company specialising in materials handling systems and logistics automation. S.A. is supported by a Royal Academy of Engineering Industrial Fellowship.}
\vspace{-2.5em}
}

\maketitle
\thispagestyle{empty}
\pagestyle{empty}

\begin{abstract}
We consider a warehouse in which dozens of mobile robots and human pickers work together to collect and deliver items within the warehouse. The fundamental problem we tackle, called the order-picking problem, is how these worker agents must coordinate their movement and actions in the warehouse to maximise performance in this task. Established industry methods using heuristic approaches require large engineering efforts to optimise for innately variable warehouse configurations. In contrast, multi-agent reinforcement learning (MARL) can be flexibly applied to diverse warehouse configurations (e.g. size, layout, number/types of workers, item replenishment frequency), and different types of order-picking paradigms (e.g. Goods-to-Person and Person-to-Goods), as the agents can learn how to cooperate optimally through experience. We develop hierarchical MARL algorithms in which a manager agent assigns goals to worker agents, and the policies of the manager and workers are co-trained toward maximising a global objective (e.g. pick rate). Our hierarchical algorithms achieve significant gains in sample efficiency over baseline MARL algorithms and overall pick rates over multiple established industry heuristics in a diverse set of warehouse configurations and different order-picking paradigms.

\end{abstract}

\section{INTRODUCTION}
\subsection{Problem Overview}
An order received by a commercial warehouse operator may comprise of several order-lines, each specifying a required item and a quantity. \textit{Order-picking} is the process of retrieving these items in the warehouse and delivering them to a target location in the warehouse for further handling~\cite{petersen1999evaluation}. The rate at which these items are retrieved in the warehouse is called the \textit{pick rate}. We describe two order-picking paradigms below: \textit{Person-to-Goods} and \textit{Goods-to-Person}.
In both paradigms, for a given set of orders, the objective is to minimise the time for order completion, which is equivalent to maximising the pick rate in expectation.

\subsubsection{\bf Person-to-Goods (PTG)}
In this paradigm, human workers will receive orders and travel around the warehouse with a push cart and pick required items manually.
We consider the augmentation of this process with robotic vehicles such as automated guided vehicles (AGVs) and autonomous mobile robots (AMRs). We primarily refer to AGVs in this work, although AMRs can be used interchangeably.
The general idea of AGV-assisted order-picking in a warehouse context has begun to receive attention in the academic literature \cite{ loffler2022picker, vzulj2022order, azadeh2023zoning}. Typically, this involves decoupling a traditional picker's role into order transportation and item picking, where transportation is handled by the AGVs and picking is handled by human or robotic pickers.
Augmentation of this paradigm with AGVs requires minimal modification to existing infrastructure and allows for scaling with variation in demand by changing AGV and picker numbers \cite{azadeh2023zoning}.

\subsubsection{\bf Goods-to-Person (GTP)}
In this paradigm, large-scale autonomous systems comprised of conveyors, picking robots and transport robots move storage mediums (such as totes, cartons or shelves) containing items to stationary human pickers, who pick and consolidate items out of the storage medium. Automation efforts have generally focused on this GTP paradigm, with numerous examples including the Dematic Multishuttle \cite{dematicmultishuttle}, Autostore \cite{dematicautostore}, Quicktron QuickBin \cite{quicktronquickbin} and Amazon KIVA \cite{Wurman_DAndrea_Mountz_2008}. In comparison to PTG systems, GTP systems have higher throughputs, but require significant capital investment and can be costlier to adjust to varying warehouse capacity and consumer demand. For these reasons, adoption is generally limited to larger operations.

\begin{figure*}[t]
\vspace{0.6em}
\centering
\includegraphics[width=\linewidth]{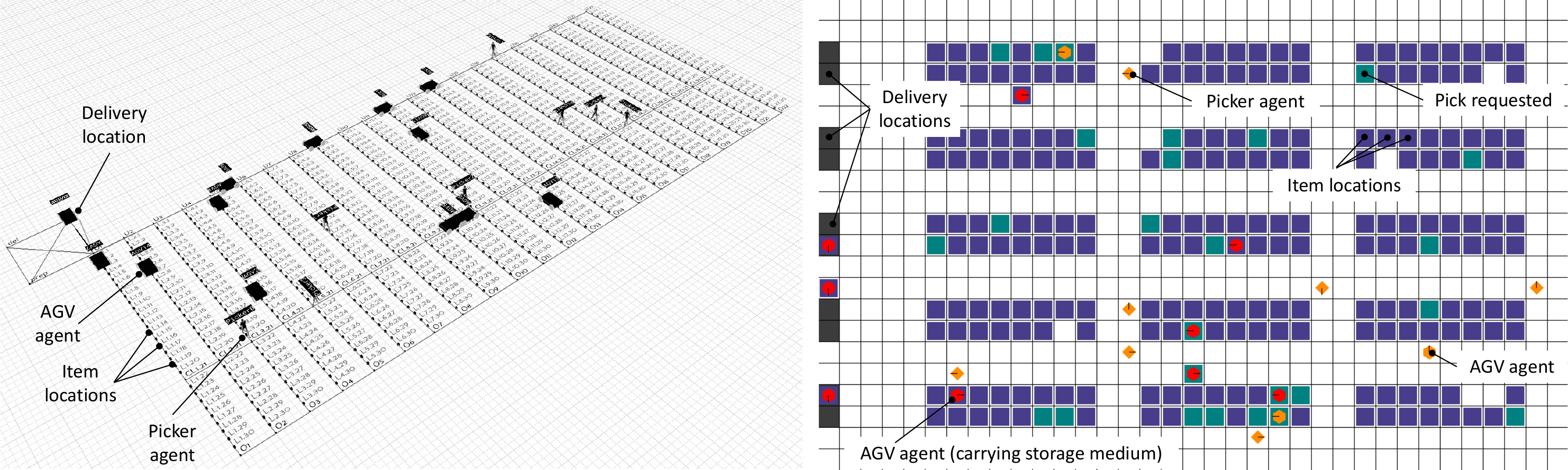}
\vspace{-1.7em}
\caption{Left: Dematic PTG simulator with human pickers and AGVs. Right: TA-RWARE GTP simulator with picking bots (diamond) and AGVs (hexagon).}
\label{fig:warehouse}
\vspace{-1.8em}
\end{figure*}

\subsection{Motivation}

Established industry methods for order-picking using heuristic approaches require significant engineering efforts to optimise for innately variable warehouse configurations \cite{azadeh2023zoning}.
Ideally, the derivation of optimal methods for worker control should be an automatic process.
Reinforcement learning (RL) offers this capability, having achieved notable successes in a
number of complex real-world domains \cite{xabi2024lyftdrivermatching, li2021huaweidynamicpickupdelivery}. Order-picking by its very nature is a multi-agent problem as it requires cooperation between multiple AGVs, robots and human pickers, thus we leverage MARL which extends RL to multi-agent systems \cite{marl-book}.
An important benefit of MARL is its flexibility to operate with diverse warehouse and worker specifications, where existing heuristic approaches require significant engineering effort and tuning to fit different specifications and stay resilient to changing factors such as demand and supply, labour conditions and order profiles.

\subsection{Contribution}

We develop a general-purpose and scalable MARL solution for the order-picking problem for warehouses with heterogeneous agents, i.e. robotic and human co-workers.
Our approach constructs a multi-layer hierarchy in which a manager agent assigns tasks to worker agents (pickers, AGVs), where each task represents a section of the warehouse (e.g. aisle) in which the worker needs to choose an item to pick. The policies of the manager and worker agents are jointly trained via MARL to maximise a global objective given by the pick rate as defined in \Cref{sec:objective}.
The hierarchical approach effectively reduces the action space of the workers by several magnitudes and facilitates better cooperation through the centrally trained manager agent.

We apply the hierarchical architecture on top of existing MARL algorithms, including Independent Actor-Critic \cite{mnih2016asynchronous, papoudakis2021benchmarking}, Shared Network Actor-Critic \cite{gupta2017cooperative}, and Shared Experience Actor-Critic \cite{christianos2020shared}, and demonstrate that it significantly improves the sample efficiency of these algorithms in a diverse set of warehouse configurations.
For our experiments, we utilise a high-performance Dematic PTG simulator which is capable of representing real-world warehouse operations.
Additionally, we introduce an open-source adaptation of a popular toy warehouse environment RWARE \cite{papoudakis2021benchmarking} to represent a GTP warehouse based on the Quicktron QuickBin systems \cite{quicktronquickbin}, named Task Assignment Multi-robot Warehouse (TA-RWARE)\footref{newrwarefootnote}. These two simulation environments are displayed in \Cref{fig:warehouse}.
We introduce competitive human-engineered heuristic methods for both picking paradigms as baselines, and
show that agents trained via our MARL algorithms achieve superior
overall pick rates in both picking paradigms.

\section{RELATED LITERATURE}
\label{sec:lit}

\subsection{AGV-Assisted Order-Picking}
\citeauthor{azadeh2023zoning} \cite{azadeh2023zoning} model the order-picking problem as a queuing network and explore the impact of different zoning strategies (no zoning and progressive zoning). They then further extend their method by representing the problem as a Markov decision process and consider dynamic switching based on the order-profile using dynamic programming. \citeauthor{loffler2022picker} \cite{loffler2022picker} consider an AGV-assisted picker and provide an exact polynomial time routing algorithm for single-block parallel-aisle warehouses. \citeauthor{vzulj2022order} \cite{vzulj2022order} consider a warehouse partitioned into disjoint picking zones, where AGVs meet pickers at handover zones to transport the orders back to the depot. They propose a heuristic for effective order-batching to reduce tardiness.
Our approach differs from these works as it does not restrict agents from accessing zones in the warehouse or constrain how the agents may collaborate with one another, leaving the discovery of effective joint strategies to the MARL algorithm. Our work aims to showcase the benefits of the inherent versatility of MARL approaches in the warehouse logistics domain, and is, to the best of our knowledge, the first application of MARL to AGV-assisted order-picking with heterogeneous agents.

\subsection{Multi-Agent Path Finding}
Reciprocal \textit{n}-body collision avoidance \cite{Berg2011ReciprocalNC}, or \textit{Multi-agent Path Finding} (MAPF), aims to build systems where teams of agents can traverse the environment to reach individually allocated targets while following optimal trajectories and avoiding collisions. MARL quickly became a promising tool for solving path-finding coordination problems, being adopted to boost scalability (PRIMAL \cite{Sartoretti2018PRIMALPV} and PRIMAL2 \cite{Damani2020PRIMAL_2PV}), enable communication \cite{gnn_MAPF} or facilitate implicit priority learning \cite{Li2022MultiAgentPF}. \textit{Lifelong-MAPF} \cite{li2021lifelong} (LMAPF) extends MAPF, as new target locations are automatically assigned to the agents upon reaching their previous goal location. \citet{li2021lifelong} utilise a centralised but bounded planner that minimises re-planning costs accumulated when receiving new targets while also showing an increase in responsiveness and adaptability of the proposed solution.
\citeauthor{greshler2021comapf} \cite{greshler2021comapf} introduce cooperative multi-agent path finding,  which is applicable within our domain but requires explicit specification of the workers that are required to cooperate and does not allow for optimisation over extended periods of time.
While we note the relevance of MAPF and especially LMAPF algorithms for the complete warehouse optimisation problem, we draw a distinction between the path-finding and order-picking settings and highlight their complementary nature.
In MAPF, task assignment processes are assumed to be external to the path planning method, in contrast to our setting where task assignment represents the main focus and path-finding is achieved through pre-defined methods.

\subsection{Multi-Agent Pickup and Delivery Problem}
Multi-agent Pickup and Delivery (MAPD) problems \cite{mapd} consider a set of agents that are sequentially assigned tasks in the form of target pickup and delivery locations, who then travel to their allocated locations while avoiding collisions with others. The objective is to minimise the time duration required for task completion, which may be further broken down into two sub-problems: Multi-Agent Task Assignment (TA) and Multi-Agent Path-Finding (MAPF) \cite{mapd,XuIROS22}.
However, MAPD approaches rely on hand-engineered heuristics that assume homogeneity among the agent architectures \cite{mapd}, with variance in agent velocities being the general extent for agent diversity \cite{Berg2011ReciprocalNC}.
This assumption is a drastic simplification of the complex coordination problem of \textit{order-picking} systems and reduces the cooperation among the agents to collision avoidance, which is solely tackled through the MAPF module.
The decoupling of workers within our approach introduces complex interdependencies between the task assignment of different worker types which significantly complicates the problem. \textit{Pickers} and \textit{AGVs} need to synchronise and meet at certain item locations at matching times in order to execute pickups. Furthermore, coordination among agents of the same type is required to minimise cramming at the same item location, which is a significant source of delays. To address the complex heterogeneous coordination problem, our approach diverges from hand-crafted heuristic-based solutions in favor of multi-agent reinforcement learning.

\subsection{Multi-Agent Reinforcement Learning}
MARL algorithms are designed to train coordinated agent policies for multiple autonomous agents, and have received much attention in recent years with the introduction of deep learning techniques into MARL \cite{marl-book,papoudakis2019dealing}.
MARL has previously seen application to various warehousing problems, including Shared Experience Actor-Critic to homogeneous GTP systems \cite{christianos2020shared, papoudakis2021benchmarking}, and a deep Q-network variant for sortation control \cite{s20123401}. For the specific complexities of the order-picking problem, we consider methods at the intersection of MARL and hierarchical RL (HRL) to enable action space decomposition and temporal abstraction. This combination has been studied by \citet{xiao2020macromarl} who derive MARL algorithms for macro-actions under partial observability, and \citet{ahilan2019feudal} who propose Feudal Multi-Agent Hierarchies (FMH) which extends Feudal RL \cite{dayan1992feudal} to the cooperative MARL domain. We introduce a 3-layer adaptation of FMH and apply it to a partially observable stochastic game with individual agent reward functions, see \Cref{sec:algo}.

\section{PRELIMINARIES}
\label{sec:prob_stat}

We consider a scenario in which a warehouse manager seeks to improve the efficiency of their warehouse, $\mathcal{W}$, through automation of order-picking. The task requires optimal utilisation of their resources to maximally improve warehouse operations, measured by the key performance indicator pick rate, defined as order-lines per hour.

\subsection{Warehouse Definition}
\label{sec:warehouse}
We define a warehouse by the 3-tuple $\mathcal{W} = \{L, Z, W\}$:
\begin{itemize}
\item $L$ refers to the set of spatially distributed locations within the warehouse, showed in Figure \ref{fig:warehouse},  and can be further broken down into $L = L_{item} \cup L_{delivery} \cup L_{other}$, where $L_{item}$ refers to the set of locations with items stored inside storage mediums, $L_{delivery}$ refers to locations where completed orders or storage mediums are delivered, and $L_{other}$ refers to other locations (e.g. idle or charging locations).
\item $Z$ defines the order distribution, which is dependent on the warehouse's supplier and customer behaviour and is assumed to be known. An order $z = \{(u_{0}, q_{0}), \dots, (u_{n}, q_{n})\}$ is sampled from $Z$. Each pair $(u_k, q_k)$ represents an order-line, where $u$ represents the item and $q$ the required quantity. Items $u$ are stored inside a storage medium at an item location $l \in L_{item}$.
\item $W = V \cup P $ represents the set of workers, where $V$ and $P$ are homogeneous sets of AGVs and pickers, respectively. AGVs $v \in V$ can visit locations $l \in L$, and pickers $p \in P$ can visit locations $l \in L_{item}$.
\end{itemize}

The order-picking paradigms we consider differ in the way items are retrieved and delivered, described below:
\begin{itemize}
\item In PTG picking, $|L_{delivery}| = 1$. In this paradigm, AGVs are assigned orders sampled from $Z$ with $z^v$ denoting the current order of AGV $v \in V$. A human picker $p \in P$ will pick an order-line $(u, q)$ out of a storage medium at an item location $l \in L_{item}$ and place it into an AGV $v$. Once the AGV has received all order-lines, the order $z^v$ is completed and the AGV will deliver it to a delivery station $l \in L_{delivery}$.
\item In GTP picking, multiple AGVs $v \in V$ will carry separate storage mediums containing items $\{{u_0,...,u_n}\}$ which are required in an order $z$. A picker robot $p \in P$ will move a storage medium from an item location $L_{item}$ containing item $u$ onto an AGV $v$, and the AGV will take the storage medium to a picking station $l \in L_{delivery}$, where an operator will pick an order-line $(u, q)$ from the AGV. Once all order-lines for order $z$ are picked, the order is completed.

\end{itemize}

\subsection{Objective}
\label{sec:objective}

For a given warehouse, we seek to derive a joint policy $\pi$ which defines the behaviour of all workers in $W$ such that $\pi$ maximises the average pick rate $K$, formally denoted with $\pi \in \argmax_{\pi} K(W,\pi)$. Pick rate is measured in completed order-lines per hour. A key desideratum of our solution is to automatically learn optimal policies for any given warehouse configuration and order profile. Specifically, we desire a general-purpose algorithm that can learn to handle variations in multiple dimensions, including the number of total item locations $|L|$, the order distribution $Z$, and the number of workers $|V| + |P|$.
Controlling all workers with a single decision-making entity becomes infeasible due to the joint action space growing exponentially with the number of workers. Hence, we consider MARL approaches in which pickers and AGVs are modeled as individual agents.

\section{PROPOSED APPROACH}
\label{sec:marl}

We introduce our proposed approach by first formulating the order-picking problem as a \textit{partially observable stochastic game} and define the observation and action spaces in \Cref{sec:prob_mod}. We then describe our hierarchical MARL approach to learn optimal agent policies  in \Cref{sec:algo}.

\subsection{Problem Modelling} \label{sec:prob_mod}
\label{sec:problem_modelling}
\subsubsection{Partially observable stochastic game}
We model the multi-agent interaction as a partially observable stochastic game (POSG) with $N$ agents \cite{hansen2004dynamic}. A POSG is defined by the tuple $(\cI, \cS, \{\cA^i\}_{i\in \cI}, \{\cO^i\}_{i\in \cI}, \cP, \Omega, \{\cR^i\}_{i\in \cI})$, with agents $i\in\cI = \{1,\ldots,N\}$, state space $\cS$, and joint action space $\cA = \cA^1\times\ldots\times \cA^N$.
At timestep $t$, each agent $i$ only perceives a partial observation $o^i_t \in \cO^i$ of the global state $s_t$ and selects an action $a^i_t$ based on action probabilities given by its policy $\pi^i(a^i_t|h^i_t)$, which in general is conditioned on the observation history $h^i_t = (o^i_1,...,o^i_t)$.
Given joint action $a_t = (a^1_t,...,a^N_t)$, the game transitions into a new state $s_{t+1}$ with probabilities given by $\cP(s_{t+1} | s_t,a_t)$, new observations are generated using probabilities given by $\Omega(o_{t+1}^1,...,o_{t+1}^N | s_{t+1}, a_t)$, and each agent $i$ receives a reward $r_t^i = \cR^i(s_t,a_t,s_{t+1})$. Agents are rewarded for behaviour aligned with the objective stated in \Cref{sec:objective}. These reward functions are environment specific and defined in \Cref{apendix_env1} and \Cref{apendix_env2}.

The goal is to learn a joint policy $\pi = (\pi^1,\ldots,\pi^N)$ to maximise the expected discounted return $G^i=\sum^T_{t=1}{\gamma^{t-1}r_t^i}$ of each agent $i$ with respect to the policies of other agents; formally, $\forall i\in\cI: \pi^i \in \argmax_{\pi^{\prime i}} \mathbb{E}\left[G^i \mid \pi^{\prime i}, \pi^{-i}\right]$ where $\pi^{-i} = \pi \setminus \{\pi^i\}$, and $\gamma$ and $T$ denoting the discount factor and episode length, respectively.

\subsubsection{Observation space}
The availability of communication links between workers and the central servers in a warehouse affords a high degree of flexibility in modelling the information observed by agents, allowing for control over the degree of partial observability. In the environments in our experiments, agents observe each other's current and target locations, and information pertinent to the order they are carrying out. Observation spaces are defined per environment in \Cref{app_env1_obs} and \Cref{app_env2_obs}.

\subsubsection{Action space}
Completion of orders requires pickers to be able to visit all item locations $l \in L_{item}$ and for AGVs to be able to visit all locations $l \in L$.
We enable this by defining the action space of pickers and AGVs as $\cA^{p} = L_{item}$ and $\cA^{v} = L$, respectively. Agents are considered \textit{busy} until they transit to their selected action location.
This proposed action space simplifies AGV and picker collaboration by allowing policies to focus on coordinating item location selection between agents while leaving the low-level navigation task to a pre-defined controller. However, this results in large action spaces that scale with the number of item locations within the warehouse, and longer durations until actions are completed. We address these issues through two techniques: (1) action masking, see \Cref{sec:vam}, and (2) the proposed hierarchical MARL architecture, detailed in \Cref{sec:algo}.

\begin{figure}
\centering
\vspace{0.6em}
\includegraphics[width=\linewidth]{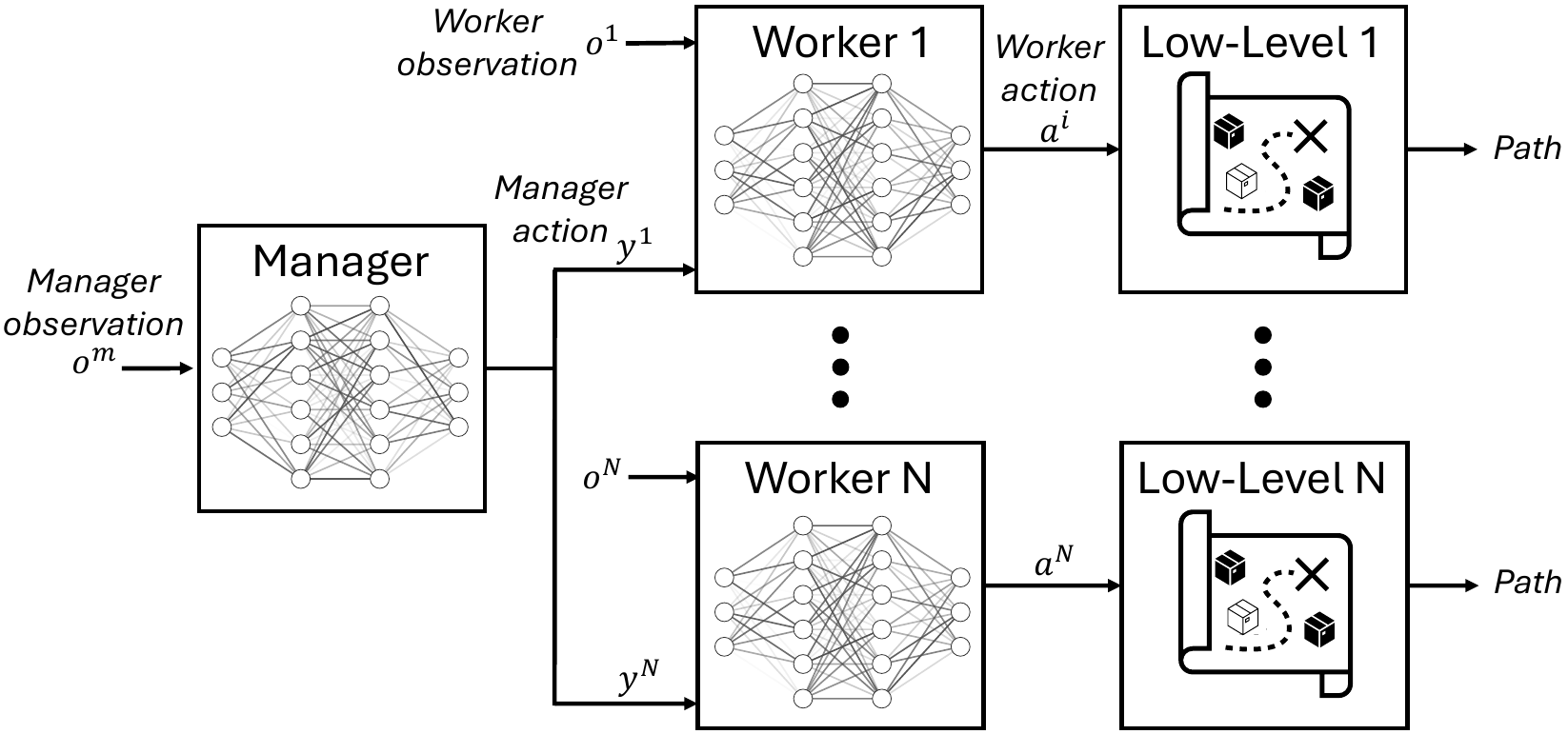}
\vspace{-1.5em}
\caption{Proposed 3-layer manager/worker agent hierarchy. A manager agent observes information about the warehouse state and orders, and assigns a task (target zone in warehouse) to each worker agent. Worker agents receive local observations about the warehouse and the assigned task from the manager, and select an item location from the assigned target zone. A low-level controller then navigates the worker to the selected item location.}
\label{fig:hier_net}
\vspace{-1.5em}
\end{figure}

\subsubsection{Invalid action masking} \label{sec:vam}

To reduce the effective action space of each agent and, thus, simplify learning, we mask out actions that would clearly be sub-optimal by adjusting logits \cite{Huang2022actionmasking}.
For instance, in PTG, one immediate observation is that whilst fulfilling an order $z$, AGVs should only move to locations $l \in L_{item}$ within the warehouse that contain requested items in $z$.
Given that in expectation, $|z| \ll |L|$, it may be advantageous for  $|\cA^v| \leq |z|$ especially when challenges related to exploration and coordination are considered. While the action-masking requires specification for these environments, the warehouse-optimisation setting we aim to tackle enables high transferability of these masks across warehouses of different sizes and types, discussed in \Cref{app_env1_vam} and \Cref{app_env2_vam}.
We note that \textit{invalid action masking} introduces bias in the generated policies which can be beneficial in early training stages but can limit the expressiveness of the policies (e.g. pickers cannot move pre-emptively to wait for AGVs). While the proposed action masking aims to minimize bias as much as possible, we leave their complete exception from our large action space training regime to future work.

\subsection{Hierarchical MARL for Order-Picking}
\label{sec:algo}

By employing a hierarchical model, we further reduce the complexity of the action space and improve the handling of the differing termination durations for actions. We introduce a 3-layer adaptation of Feudal Multi-Agent Hierarchies (FMH) \cite{ahilan2019feudal}, which involves the introduction of a manager agent that produces goals for worker agents to satisfy, shown in \Cref{fig:hier_net}. In contrast to FMH, manager goals do not affect worker reward functions, but instead the goals partition the worker action spaces (as defined below), and worker agents do not execute primitive actions in the environment, instead delegating their decisions to lower-level controllers.
The manager goals divide the locations within the warehouse into a set of disjoint zones $Y$, formally $L = \bigcup_{y\in Y}y$.
The manager's action space consists of a choice from the set of zones $Y$ for each agent $i \in \cI$, given by $\cA^{m} = Y^{|\cI|}$.
Given assigned zone $y^i$ to worker agent $i$, its policy $\pi^i$ selects a new target location $l_t^i \in y^i$ within the assigned zone. This decomposition greatly reduces the effective action space of each agent's policy, which is now bounded by $\max_{y\in Y} |y| \ll |L|$. Once the target location of each worker agent is determined, a lower-level controller will calculate the shortest path from its current location and execute the necessary sequence of primitive actions (we use the A* algorithm \cite{Hart1968astar} in our experiments).

The manager reward $r^m_t$ is the sum of the rewards from all the assigned goals to non-busy workers $i \in \cI$ during a timestep.
This is equivalent to the sum of rewards received by the non-busy workers after each intermediate step $\tau \in [t,t+k_i]$, where $k_i$ represents the number of steps taken by worker $i$ before reaching the goal:
\begin{align}
\vspace{-1em}
\label{eq:mng_rew2}
r^m_t &= \sum_{i \in \cI} r^i_{t:t+k_i} \text{, with} \\
r^i_{t:t+k_i} &=
\begin{cases}
\sum_{\tau=t}^{t+k_i} r^i_\tau, & \text{if $i$ received a goal at } t\\
0, & \text{otherwise}
\end{cases}
\end{align}

The manager and worker policies are trained jointly via MARL. In this work, we analyse performance improvements across different data-sharing mechanisms for MARL:

\begin{itemize}
\item \textbf{Independent Actor-Critic (IAC)} \cite{mnih2016asynchronous, papoudakis2021benchmarking} -- each picker and AGV have their own independent networks, allowing for specialised behaviours, but no shared experience between agents of the same type.
\item \textbf{Shared Network Actor-Critic (SNAC)} \cite{gupta2017cooperative, christianos2021scaling} -- networks are shared across pickers and AGVs respectively, to improve the efficiency of the training process.
\item \textbf{Shared Experience Actor-Critic (SEAC)} \cite{christianos2020shared} -- each picker and AGV have their own independent networks, but have an additional shared gradient update across agents of the same agent type.
\end{itemize}
We refer to the hierarchical versions of these algorithms as HIAC, HSNAC, and HSEAC.

\section{EMPIRICAL EVALUATION}
\label{sec:sim}

\subsection{Warehouse Simulators}
To test the generality of our algorithms across different warehouse layouts and picking paradigms, we utilise two simulator environments, described below.
\subsubsection{\bf{Dematic PTG Simulator}}
Our first environment is Dematic's high-performance PTG warehouse simulator, which is capable of representing real-world warehouses and PTG picking operations. An example snapshot of a simulated warehouse in our experiments is shown in \Cref{fig:warehouse} (left). Pickers in this simulator are human workers, and cooperate with AGVs that transport picked items. AGVs travel to multiple storage locations to receive items successively, picked by human pickers. Once all items within an order are collected, the AGV delivers the order to a single delivery location.

\subsubsection{\bf{TA-RWARE}} Our second environment is an open-source simulator named TA-RWARE\footnote{\scriptsize\label{newrwarefootnote}\url{https://github.com/uoe-agents/task-assignment-robotic-warehouse}}, an extension of the popular toy warehouse environment RWARE \cite{papoudakis2021benchmarking} tailored towards the GTP paradigm, as shown in \Cref{fig:warehouse} (right). In order to create a cooperative task and study task assignment optimisation, we designed TA-RWARE to include heterogeneous agents (AGVs and picker robots) which select target locations as actions, with map traversal from one location to another being handled by a predefined heuristic.
In this environment, AGVs travel to a single warehouse location to retrieve a storage medium containing items, transferred to the AGV by a picking robot. The storage mediums are then delivered to one of multiple delivery locations which is a human pick station. The human picker in this system is not modelled in our simulator, and sits outside the bounds of the system at the delivery locations. As such, when we refer to a picker in the context of this simulator, we are referring to a picking robot which lifts storage mediums (such as totes or boxes) onto the AGV.

\begin{table*}[h]
\vspace{0.6em}
\caption{Performance comparisons between heuristics and marl algorithms, showing mean $\pm$ 95\% ci pick rate in order-lines per hour.}
\centering
\vspace{-0.6em}
\begin{tabular}{l|cccc|cc|}
\cline{2-7}
& \multicolumn{4}{c|}{Dematic Simulator (PTG) Environment}                                                                             & \multicolumn{2}{c|}{TA-RWARE (GTP) Environment}      \\
\multicolumn{1}{c|}{}       & \multicolumn{1}{c}{Small}     & \multicolumn{1}{c}{Medium}    & \multicolumn{1}{c}{Large}     & \multicolumn{1}{c|}{Disjoint} & \multicolumn{1}{c}{Small}   & \multicolumn{1}{c|}{Large} \\ \hline
\multicolumn{1}{|l|}{FM}    & $901.3 \pm 1.9$               & $1098.1 \pm 3.8$              & $1230.2 \pm 5.1$              & $568.4 \pm 1.7$               & \multicolumn{1}{c}{--}      & \multicolumn{1}{c|}{--}     \\
\multicolumn{1}{|l|}{PDM}   & $783.6 \pm 2.8$               & $982.2 \pm 4.0$               & $1123.9 \pm 4.9$              & $677.4 \pm 2.1$               & \multicolumn{1}{c}{--}      & \multicolumn{1}{c|}{--}     \\
\multicolumn{1}{|l|}{CTA}   & \multicolumn{1}{c}{--}        & \multicolumn{1}{c}{--}        & \multicolumn{1}{c}{--}        & \multicolumn{1}{c|}{--}       & $52.7 \pm 0.9$              & $67.1 \pm 0.8$             \\
\multicolumn{1}{|l|}{IAC}   & $1053.0 \pm 2.8$              & $1206.4 \pm 4.2$              & $1263.9 \pm 5.8$              & $733.2 \pm 2.7$               & $65.2 \pm 0.5$              & $80.4 \pm 0.6$           \\
\multicolumn{1}{|l|}{SNAC}  & $990.9 \pm 2.8$               & $1142.7 \pm 4.3$              & $1235.0 \pm 5.7$              & $688.7 \pm 2.7$               & $60.8 \pm 0.7$              & $72.1 \pm 0.9$             \\
\multicolumn{1}{|l|}{SEAC}  & $1019.7 \pm 2.9$              & $1185.1 \pm 5.1$              & $1262.9 \pm 5.7$              & $739.8 \pm 2.4$               & $64.8 \pm 0.4$              & $82.2 \pm 0.5$             \\
\multicolumn{1}{|l|}{HIAC (ours)}  & $1025.9 \pm 4.3$              & $1232.1 \pm 4.8$              & $1354.2 \pm 5.9$              & $794.1 \pm 2.7$            & $66.7 \pm 0.3$          & $86.0 \pm 0.5$        \\
\multicolumn{1}{|l|}{HSNAC (ours)} & $1030.8 \pm 3.8$              & $1232.8 \pm 5.1$              & $1363.8 \pm 6.0$              & $796.9 \pm 2.4$            & $66.0 \pm 0.7$          & $85.0 \pm 0.5$     \\
\multicolumn{1}{|l|}{HSEAC (ours)} & $1028.2 \pm 3.9$              & $1242.1 \pm 5.0$              & $1370.9 \pm 5.7$              & $803.5 \pm 2.6$            & $64.6 \pm 0.4$          & $84.8 \pm 0.6$           \\ \hline
\end{tabular}
\label{tab:algorithm_comparison}
\vspace{-2em}
\end{table*}

\subsection{Heuristic Solutions}
\label{sec:heuristics}

Two established industry heuristics used by Dematic for order-picking under the PTG paradigm are \textit{Follow Me} (FM) and \textit{Pick, Don't Move} (PDM) (these are similar to the strategies described by Azadeh et al.\cite{azadeh2023zoning} as No Zoning and Progressive Zoning). We define a third heuristic for the GTP paradigm, which we call \textit{Closest Task Assignment} (CTA).
\subsubsection{\bf Follow Me (FM)}
Multiple AGVs are assigned to each picker (i.e. they form a group) and will follow them through the warehouse. Each AGV's order is concatenated and the travelling salesman problem (TSP) solution is generated to determine the order in which the items will be picked. The TSP path minimises the distance of each group of workers with the constraint that they stay together while orders are not completed.
FM minimises idle time for pickers, as it ensures that they are always travelling or picking, but can also lead to more travelling of pickers than needed.
\subsubsection{\bf Pick, Don't Move (PDM)}
Pickers are allocated to zones (e.g. a picker per aisle) in the warehouse which they are responsible for, while AGVs are allowed to travel throughout the entirety of the warehouse. The AGVs travel to all item locations in their current order using a TSP solution.
Pickers meet AGVs that travel into the picker's designated zones at the required item location and pick items into the AGV.
Pickers prioritise service of AGVs by the relative proximity of the AGV and picker to the target locations.
PDM minimises travel distance for pickers, however, it may result in under-utilisation of pickers in case there are few items within current orders in their operating zones.

\subsubsection{\bf Closest Task Assignment (CTA)}
AGVs travel to single storage locations and deliver storage mediums from those storage locations to a plurality of delivery locations. Storage mediums that need to be picked are assigned to the closest AGV, which takes the storage medium to the closest delivery location. Once delivered, the AGV then returns the storage medium to the closest empty shelf location. \textit{Closest} in this context refers to the minimum distance path found by the A* algorithm \cite{Hart1968astar}. Pickers stick to allocated zones (similar to PDM), but prioritise AGVs in a first-in-first-out (FIFO) queue according to which AGV was assigned a pick or a drop within its zone first. Similarly to PDM, CTA minimises travel distance for pickers but may also result in picker under-utilisation.
\subsection{Experiments}

We evaluate the algorithms in four different PTG environment configurations and two different GTP environment configurations based on Dematic customer warehouse profiles.
Full specifications for the PTG and GTP configurations are provided in \Cref{app_env1_config}  and \Cref{app_env2_config}, respectively.
We compare all three hierarchical algorithms, HIAC, HSNAC, HSEAC against the PDM, FM and CTA heuristics (\Cref{sec:heuristics}) as well as the non-hierarchical baselines, IAC, SNAC and SEAC, in both the PTG and GTP environments. The non-hierarchical baselines use the same neural network architecture as the worker agents in their respective hierarchical versions. Details on architecture and hyperparameter values are given in \Cref{apendix_parameters}. We use pick rate measured in order-lines per hour as our primary performance measure, indicating the average frequency of picks in each episode.

\begin{figure}
\vspace{0.5em}
\hspace{-0.6em}
\centering
\includegraphics[width=1.01\linewidth]{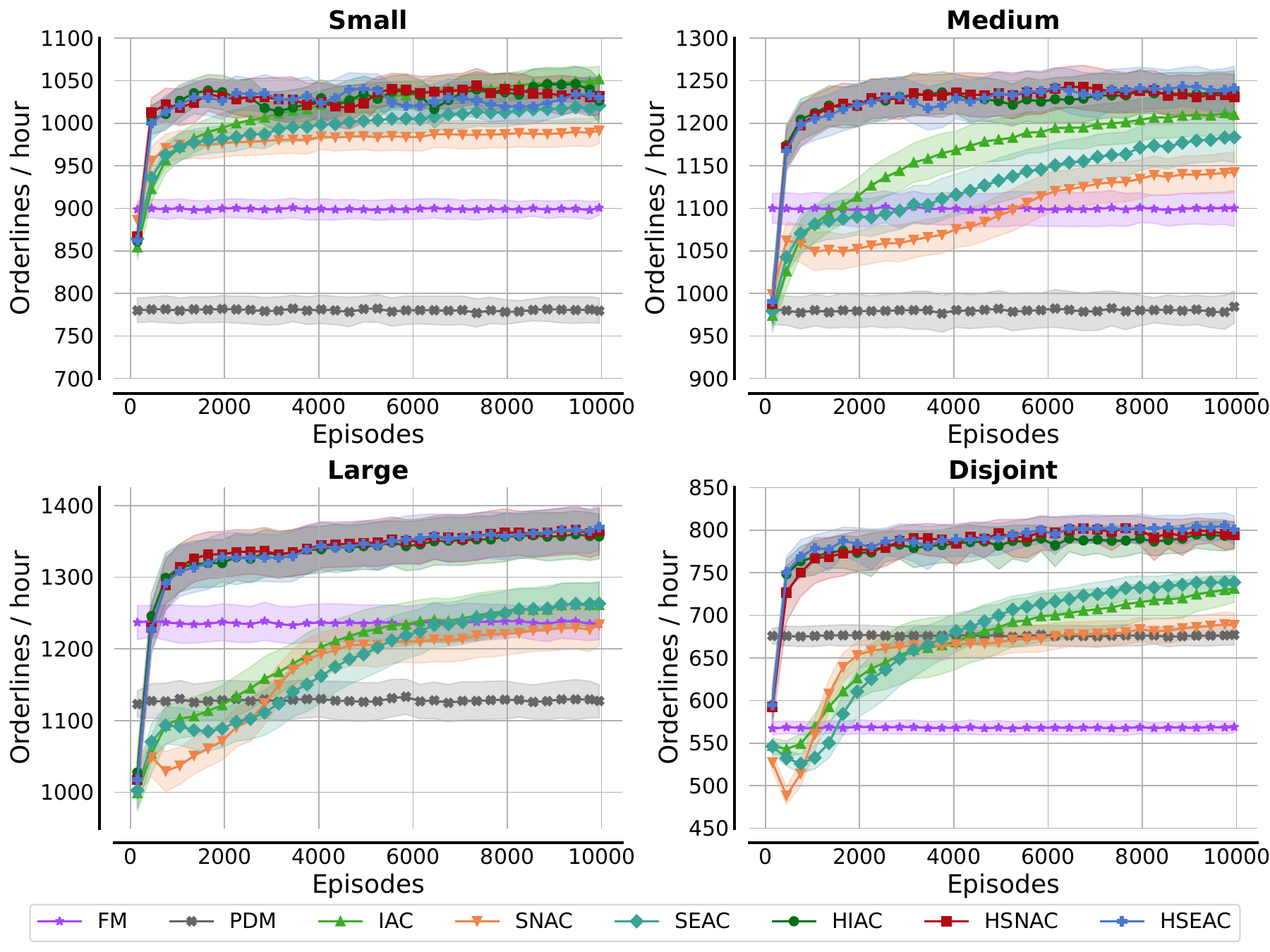}
\vspace{-1.6em}
\caption{Average pick rate (order-lines per hour) in Dematic PTG simulator for heuristics FM/PDM and MARL algorithms IAC, SNAC, SEAC, HIAC (ours), HSNAC (ours), HSEAC (ours). Shaded area shows 95\% stratified bootstrap confidence interval \cite{agarwal2021deep}, with 300 episode average smoothing.}
\label{fig:main_results}
\vspace{-1.75em}
\end{figure}

\begin{figure}
\vspace{1em}
\hspace{0.1em}
\includegraphics[width=1\linewidth]{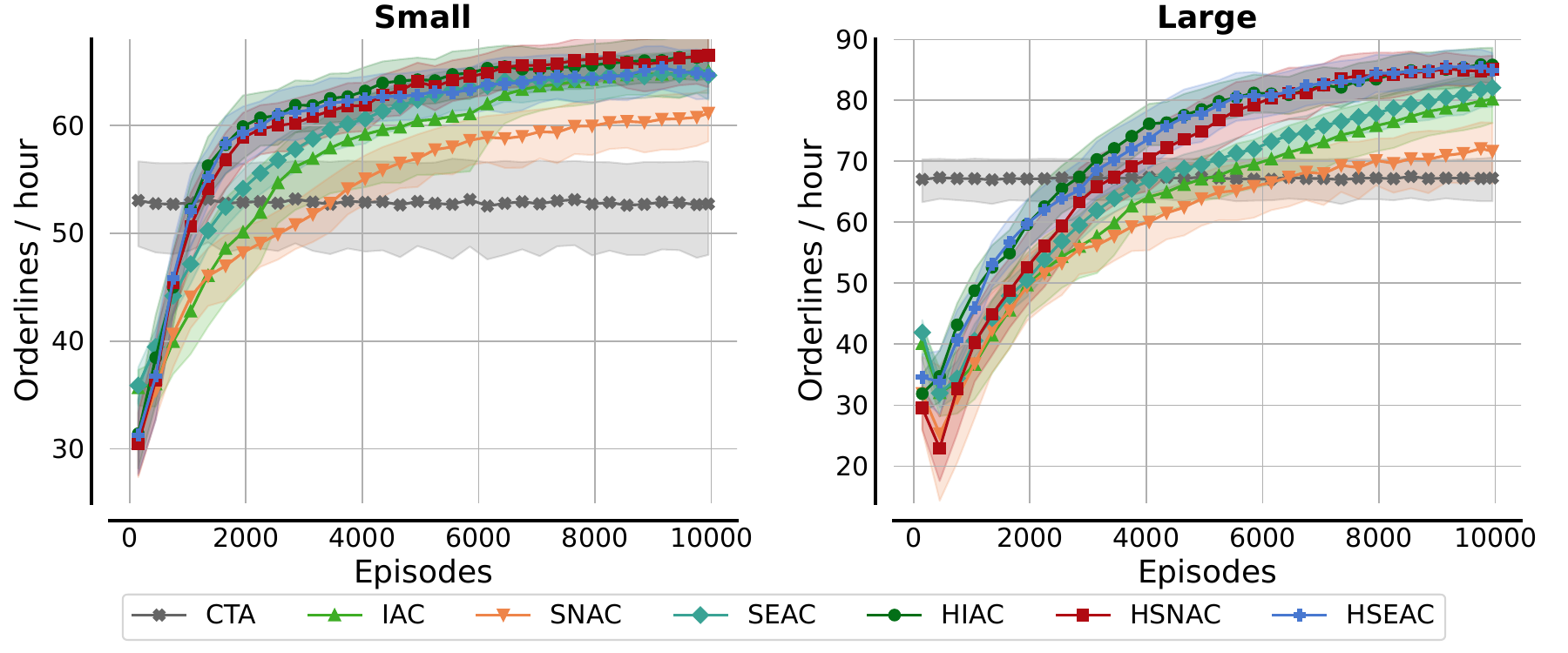}
\vspace{-1.6em}
\caption{Average pick rate (order-lines per hour) in TA-RWARE GTP simulator for heuristic CTA and MARL algorithms IAC, SNAC, SEAC, HIAC (ours), HSNAC (ours), HSEAC (ours). Shaded area shows 95\% stratified bootstrap confidence interval, with 300 episode average smoothing.}
\label{fig:main_results_rware}
\vspace{-1.75em}
\end{figure}

\Cref{fig:main_results} shows the pick rate for HIAC, HSNAC, HSEAC and all baselines across training in PTG.
While different warehouse configurations can favour one heuristic or the other (e.g. FM in Large, PDM in Disjoint), we observe that the hierarchical algorithms achieve significantly higher pick rates than these two heuristics independent of the warehouse setting.
Comparing the hierarchical versions against the original algorithms (i.e. HIAC to IAC, HSNAC to SNAC, HSEAC to SEAC) demonstrates the advantage of the hierarchical architecture, with all hierarchical versions showcasing superior sample efficiency, especially as the complexity of the warehouse increases.
We perform a similar analysis for GTP, shown in \Cref{fig:main_results_rware}.
Contrasting the pick rates achieved by the proposed methods against the CTA heuristic and the MARL algorithms,
we note that all MARL methods surpass the heuristic in both settings, with SNAC achieving the lowest pick rates by the end of training.
The sub-par performance of SNAC is due to all worker agents using identical policies, which can lead to delays incurred due to frequent collisions or deadlocks among workers (this limitation of SNAC was also observed in prior work \cite{christianos2020shared}).
The advantage of the hierarchical architecture can be observed when comparing HSNAC to SNAC
. HSNAC avoids deadlocks through the manager,
by conditioning the worker policies on assigned target goals to distribute them across the warehouse.
Finally, analogous to the PTG environment, the sample efficiency is again superior for the hierarchical models, especially in the Large configuration, highlighting again the scaling benefits when compared to the non-hierarchical baselines.

In \Cref{tab:algorithm_comparison}, we compare the average pick rates achieved by the algorithms during the final $50$ training episodes.
In the Small PTG configuration, IAC achieves on-par pick rates with the hierarchical algorithms, a difference of only $2.2\%$, while surpassing FM and PDM by $16.8\%$ and $34.4\%$ respectively. We attribute these results to the relatively low difficulty of the task where the hierarchical approach does not offer substantial benefits. As we scale up warehouse complexity, the hierarchical algorithms start reaching the highest overall pick rates. In Medium, HSEAC exceeds FM and PDM by $13.1\%$ and $26.5\%$. In Large, HSEAC exceeds FM and PDM by $11.4\%$ and $22.0\%$. In Disjoint, HSEAC exceeds FM and PDM by $41.3\%$ and $18.6\%$. In GTP, the hierarchical models achieve the highest pick rates, HIAC surpassing the CTA heuristic by $26.6\%$ in the Small configuration and by $28.2\%$ in the Large configuration.

\section{CONCLUSION}
\label{sec:discuss}
Our results support our hypothesis that MARL algorithms can derive effective solutions for the order-picking problem in both PTG and GTP picking paradigms.
We constructed a hierarchical MARL architecture consisting of a manager agent that assigns individual goals to different groups of worker agents inside the warehouse, and the policies of the manager and worker agents are jointly trained using MARL.
Our solution builds on top of and outperforms several MARL baselines that integrate different experience-sharing mechanisms.
We attribute the performance improvement over the baselines to the hierarchical decomposition of large action spaces, allowing for a solution to the order-picking problem at a lower spatial resolution. The hierarchical approach also provides a high-level central coordination mechanism, as goals for all agents are selected by a single manager policy.
The proposed MARL solutions outperform multiple engineered and well-established industry heuristics in various warehouse configurations across both PTG and GTP paradigms.
In future work, we intend to explore the inclusion of other optimisation objectives into our objective function. Measures such as travel distance and energy usage are often of high importance to warehouse managers, as they have real-world ramifications in terms of maintenance costs, operational costs, and human employee welfare.
To further facilitate efficient scaling in number of agents and warehouse size, methods based on unsupervised environment design \cite{garcin2024dred} and sub-task decomposition \cite{fosongLearningComplexTeamwork2024} could be developed.

\appendix

\subsection{Hyperparameters \& Training Configuration} \label{apendix_parameters}
The manager policy and value network are multi-headed neural networks comprising of three fully-connected layers of 128 neurons each with ReLU activations. Each agent is parameterised by a value and critic network represented by two fully connected layers of 64 neurons with ReLU activations. We use the same algorithm hyperparameter values in all MARL algorithms and experiments: learning rate is 0.0003, network update frequency is 100 steps for PTG and 250 steps for GTP, Adam optimiser epsilon is 0.001, and GAE lamba parameter is 0.96. The discount factor is $0.99$ for all agents.
In all our experiments we train for $10,000$ episodes.
The partitioning, $Y$, is achieved through the division of the warehouse into equal-sized sections depending on the configuration, shown in tables in \Cref{app_env1_config,app_env2_config}.

\subsection{Environment 1 --- Dematic Simulator (PTG)}  \label{apendix_env1}
\subsubsection{Order-picking dynamics}
The agents are presented with an episodic task consisting of $N$ orders that are randomly distributed within locations $L_{item}$, terminating when all orders are completed.

\subsubsection{Warehouse configurations} \hfill \label{app_env1_config}

\begin{table}[H]
\vspace{-0.75em}
\begin{center}
\begin{tabular}{@{}l@{}cccc@{}}

\toprule
& Small & Medium & Large & Disjoint  \tablefootnote{The Disjoint warehouse is separated into two sub-warehouses (for e.g. regular and frozen goods) joined by a passage.} \\ \midrule
Aisles                           & 2     & 10     & 22    & 12 + 12  \\
Item Locations $|L_{item}|$      & 200   & 400    & 1276  & 1392     \\
Partitions $|Y|$                 & 4     & 10     & 22    & 24       \\
Pickers $|P|$                    & 4     & 6      & 8     & 4        \\
AGVs $|V|$                       & 8     & 12     & 16    & 16       \\
Avg. order-lines per order $\mathbb{E}(|z^v|)$      & 5     & 5      & 5     & 2        \\
Orders $|Z|$                     & 80    & 80     & 80    & 80       \\ \bottomrule
\end{tabular}
\label{tab:warehouse_config_comparison}
\end{center}
\vspace{-1.75em}
\end{table}

\subsubsection{Observation space} \label{app_env1_obs}
The Manager, picker and AGV observations are defined in \Cref{eq:manager_obs,eq:pick_obs,eq:agv_obs}, respectively, with $\oplus$ denoting the concatenation operator:
\begin{align}
\label{eq:manager_obs}
O^m &= \{(l_c^i, l_t^i) \mid i \in \cI\} \oplus \{z^v \mid v \in V\} \\
\label{eq:pick_obs}
O^p &= \{(l_c^i, l_t^i) \mid i \in \cI\} \oplus \{z^v \mid v \in V\} \\
\label{eq:agv_obs}
O^v &= \{(l_c^i, l_t^i) \mid i \in \cI\} \oplus z^v
\end{align}

The manager, pickers and AGVs observe the current and target locations of all agents, denoted $l_{c}^i \in L$ and $l_t^i \in L$ for agent $i$. Additionally, the manager and pickers observe all orders $z^v, v \in V$ while AGVs only observe their own order.

\subsubsection{Reward function}
Pickers are rewarded +0.1 for picking an item onto an AGV. AGVs are rewarded +0.1 for receiving a picked item, and +0.1 for delivering the order. Both agent types receive a fixed -0.01 penalty per timestep.

\subsubsection{Invalid action masking} \label{app_env1_vam}
Action masking is order-specific for AGVs,  item locations that are not part of the current order are removed from the action space. Pickers'  invalid action mask enables them to choose between the target item locations of AGVs to favour coordination. Lastly, pickers cannot choose locations that others are in transit to.

\subsection {Environment 2 --- TA-RWARE (GTP)}  \label{apendix_env2}
\subsubsection{Order-picking dynamics}
The agents choose to pick a storage medium from a dynamic request queue of fixed length (based on the warehouse layout), where a new storage medium becomes requested upon the delivery of another.

\subsubsection{Warehouse configurations} \hfill \label{app_env2_config}
\begin{table}[h]
\vspace{-0.75em}
\begin{center}
\begin{tabular}{@{}l@{}cccc@{}}

\toprule
& Small & Large  \\ \midrule
Rack Rows                           & 2      & 4      \\
Rack Columns                        & 5      & 7      \\
Column Length                       & 8      & 8      \\
Column Width                        & 2      & 2      \\
Item Locations $|L_{item}|$         & 160    & 448    \\
Partitions $|Y|$                    & 10     & 28     \\
Pickers $|P|$                       & 4      & 7      \\
AGVs $|V|$                          & 8      & 14     \\
Concurrent requested items          & 20     & 60       \\
Delivery Locations $|L_{delivery}|$ & 10     & 14      \\ \bottomrule

\end{tabular}
\label{tab:warehouse_config_comparison2}
\end{center}
\vspace{-1.7em}
\end{table}

\subsubsection{Observation space} \label{app_env2_obs}
The Manager, picker and AGV observations are defined in \Cref{eq:agent_mng_rware,eq:agent_picker_rware,eq:agent_agv_rware}, respectively:
\begin{align}
\hspace{0.5em}
\begin{split}
\label{eq:agent_mng_rware}
O^m &= \{(l_c^i, l_t^i) \mid i \in \cI\} \oplus \{(cr_v, re_v, ld_v) \mid v \in V\} \\
& \oplus \{{oc_l, re_l}\} \mid l \in L_{item}\} \\
\end{split}
\end{align}
\vspace{-1.25em}
\begin{align}
\begin{split}
\label{eq:agent_agv_rware}
O_p &= \{(l_c^i, l_t^i) \mid i \in \cI\} \oplus \{(cr_v, re_v, ld_v) \mid v \in V\} \\
\end{split}
\end{align}
\vspace{-1.5em}
\begin{align}
\hspace{-0.65em}
\begin{split}
\label{eq:agent_picker_rware}
O_v &=  \{(l_c^i, l_t^i) \mid i \in \cI\} \oplus (cr_{own}, re_{own}, ld_{own}) \\
& \oplus \{{oc_l, re_l}\} \mid l \in L_{item}\} \\
\end{split}
\end{align}
The manager, pickers and AGVs observe the current and target locations of all agents denoted $l_{c}^i \in L$ and $l_t^i \in L$ for agent $i$. Additionally, the manager and pickers observe all AGV statuses: carrying shelf, $cr_v$, carried shelf is requested, $re_v$, and waiting for load/unload, $ld_v$. AGVs only observe their own status $(cr_{own}, re_{own}, ld_{own})$. The manager and the AGVs also observe the statuses of all shelf locations $l \in L_{item}$, occupied by shelf, $oc_l$, and shelf requested state, $re_l$.
\subsubsection{Reward function}
Pickers are rewarded +0.1 for loading/unloading a storage medium onto an AGV. AGVs are rewarded +1 for delivering the storage medium. Both agent types receive a fixed -0.001 penalty per timestep.

\subsubsection{Invalid action masking} \label{app_env2_vam}
Action masking for AGVs reduces their practical action space to the shared pool of requested storage medium locations and delivery locations. For pickers, we follow the same masking scheme as in the PTG paradigm, where pickers can travel to load/unload from current AGV target locations that are not already serviced.

\bibliography{iros2024}

\end{document}